\documentclass[runningheads]{llncs}
\usepackage[T1]{fontenc}
\usepackage{graphicx}

\usepackage{bbm}
\usepackage{cite}
\usepackage{amsmath,amssymb,amsfonts}
\usepackage{algorithmic}
\usepackage{textcomp}
\usepackage{xcolor}
\usepackage{hyperref}
\usepackage{cleveref}
\usepackage{multirow}
\usepackage{booktabs}
\usepackage{subcaption}
\usepackage{caption}
\usepackage{tabularx}
\usepackage{comment}

\usepackage{array} %
\newcolumntype{Y}{>{\centering\arraybackslash}X}

\begin{document}
\newcommand{\red}[1]{{\textcolor{black}{#1}}}
\newcommand{\blue}[1]{{\textcolor{black}{#1}}}
\newcommand{\cread}[1]{{\textcolor{black}{#1}}}

\title{Accelerated Decoding of Centroid Positional Encoding for Instance Segmentation}
\titlerunning{Accelerated Decoding of Centroid Positional Encoding}

\author{Carmelo Scribano\inst{1}\and Filippo Muzzini\inst{1} \and Nedyalko Prisadnikov\inst{2}  Mohammad Mahdi\inst{2} \and Yuqian Fu\inst{2} \and Giorgia Franchini\inst{1} \and Danda Pani Paudel\inst{2} \and Marko Bertogna\inst{1} \and Luc Van Gool\inst{2}}

\authorrunning{C. Scribano et al.}
\institute{
University of Modena and Reggio Emilia, Italy\\
\email{\{name\}.\{surname\}@unimore.it}\and
INSAIT, Sofia University ``St. Kliment Ohridski'', Bulgaria
\email{\{name\}.\{surname\}@insait.ai}}

\maketitle              %

\begin{abstract}
Beyond model inference, the decoding stage, which converts raw network outputs into task-level representations, constitutes a significant portion of the execution cost. Despite its practical impact, prediction decoding has received comparatively little attention and is often implemented using generic CPU routines or inefficient GPU kernels, limiting the benefits of advances in model efficiency. In this work, we investigate the decoding overhead associated with a recent sinusoidal centroid encoding for Instance Segmentation, in which each pixel regresses a positional embedding of its instance centroid. This approach allows flexible segmentation without predefined proposals, but extracting instance masks from dense embeddings incurs a high computational cost. We present an optimized CUDA-based implementation of the decoding algorithm tailored to this encoding, explicitly addressing challenges related to parallelization, synchronization, and memory access on modern GPUs. Our solution significantly reduces decoding overhead and improves End-to-End inference latency, outperforming both CPU-based approaches and naive GPU implementations. The results demonstrate that efficient decoding is essential to fully exploit the advantages of advanced output representations and highlight the importance of jointly designing encoding schemes and their decoding algorithms for real-time computer vision systems.

\end{abstract}

\section{Introduction}

The computational cost of modern computer vision systems has been steadily increasing, along with the performance achieved on a broad set of tasks. While significant effort has been devoted to designing more efficient model architectures and advanced model compression strategies, the computational cost of prediction decoding (the transformation from raw network outputs to task-level representations) has received significantly less interest. In practice, the decoding stage can incur significant overhead and, in some cases, dominate end-to-end inference time. This issue is particularly relevant in edge deployments, where computational resources are limited and real-time processing is often required. For example, smart-city applications frequently execute vision models directly on edge devices~\cite{scribano2025edge}, making low-latency inference essential for timely decision-making. Common instances of decoding operations include non-maximum suppression in object detection~\cite{Hosang_2017_CVPR}, clustering-based grouping in instance segmentation~\cite{Jiang_2020_CVPR}, and parts grouping in human pose estimation~\cite{Tang_2019_CVPR}. These decoding steps are often implemented using generic CPU routines or suboptimal GPU kernels, leading to memory-bound operations, excessive synchronization, and poor utilization of modern accelerators. As a result, improvements in model efficiency or backbone speed do not necessarily translate into proportional gains in overall system performance.\\

\red{This study explores the computational efficiency of sinusoidal centroid encodings for instance segmentation~\cite{prisadnikov2024simple}. Although this representation enables the modeling of an arbitrary number of instances without complex proposal logic, recovering the final masks remains a bottleneck due to the multi-stage nature of the decoding algorithm. We present a high-performance CUDA implementation specifically designed to overcome these parallelization and memory access challenges. Our optimized decoder enables the model to achieve its full potential in real-time scenarios, effectively bridging the gap between theoretical representation and practical deployment. Our findings highlight that the practical value of novel output representations is closely linked to the efficiency of their associated decoding algorithms.}

\section{Background and Related Work}

\subsection{Instance Segmentation}
Formally, the goal of the instance segmentation (IS) task is to assign each pixel in an image to a unique object instance. This task is strictly related to Semantic Segmentation (SS), which instead assigns each pixel to a semantic class. The two tasks can be combined as Panoptic Segmentation, which assigns pixels to instance masks with a semantic label. This paper focuses on the class-agnostic IS task.

\begin{figure}
    \centering
    \begin{subfigure}[t]{0.24\textwidth}
        \centering
        \includegraphics[width=\textwidth,clip]{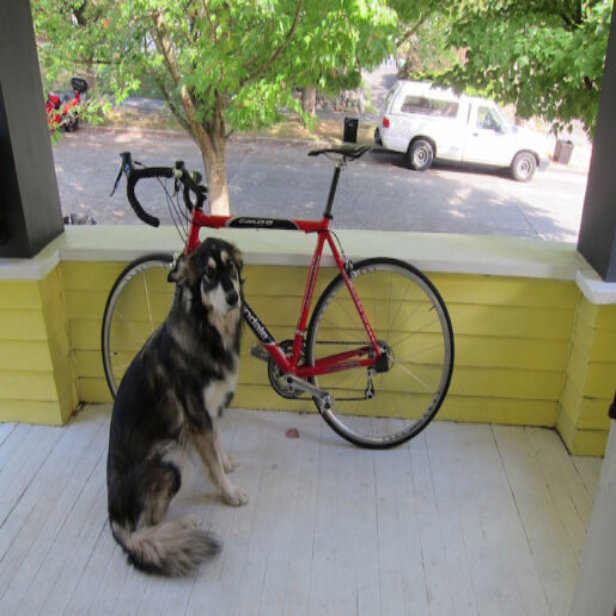}
        \caption{Source}
        \label{fig:decode_a}
    \end{subfigure}
    \begin{subfigure}[t]{0.24\textwidth}
        \centering
        \includegraphics[width=\columnwidth,clip]{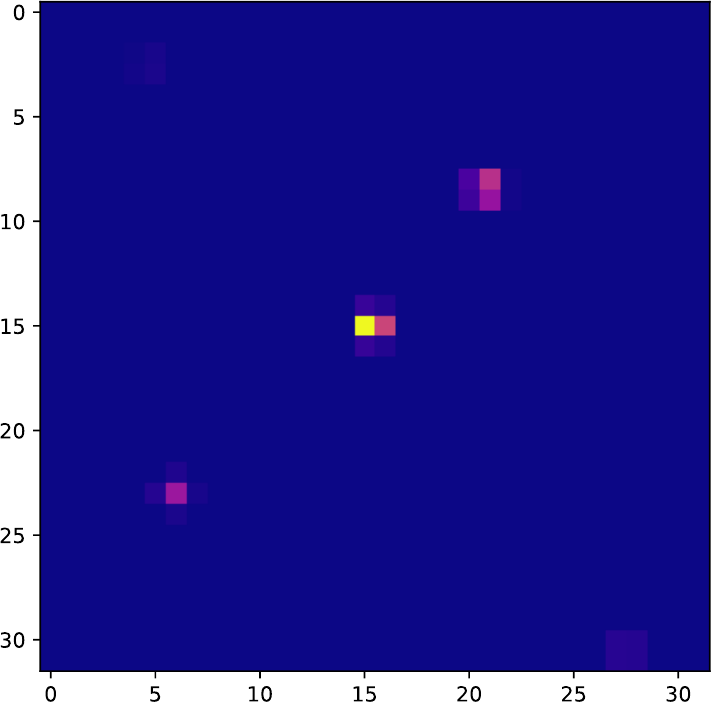}
        \caption{Votes Histogram}
        \label{fig:decode_b}
    \end{subfigure}
        \begin{subfigure}[t]{0.24\textwidth}
        \centering
        \includegraphics[width=\columnwidth,clip]{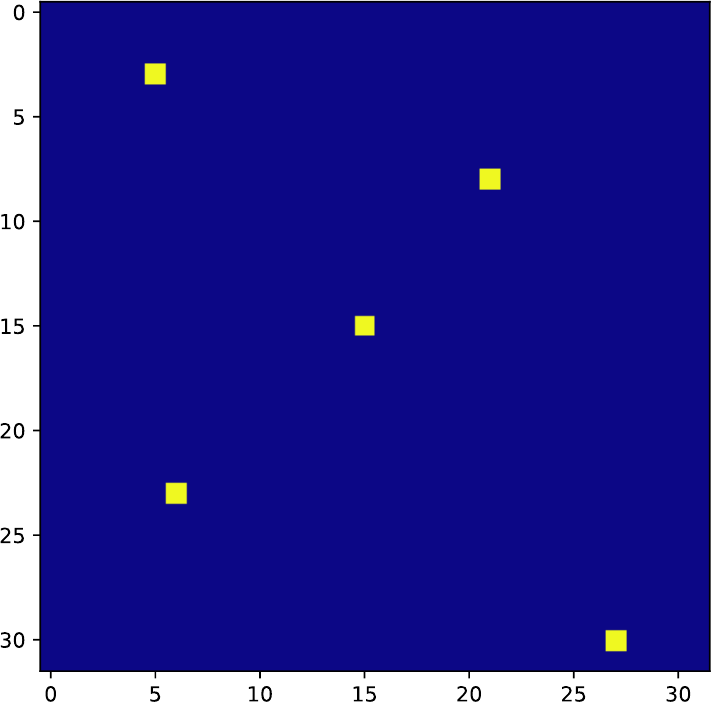}
        \caption{LocalMax Indicator}
        \label{fig:decode_c}
    \end{subfigure}
    \begin{subfigure}[t]{0.24\textwidth}
        \centering
        \includegraphics[width=\textwidth,clip]{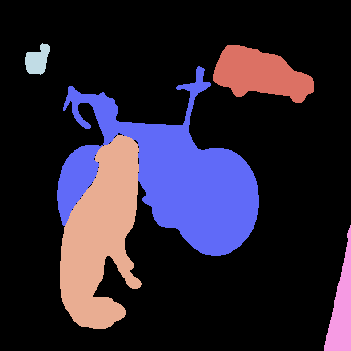}
        \caption{Instance Masks}
        \label{fig:decode_d}
    \end{subfigure}
    \caption{Overview of the decoding process for the proposed encoding.}
    \label{fig:decode}
\end{figure}

\red{Instance Segmentation (IS) is challenging because the number of object instances is not known in advance. As a result, the model must simultaneously localize objects and assign pixels to the correct instance, producing a variable number of output masks. \textbf{Top-Down} methods (e.g., Mask R-CNN \cite{he2017mask}) simplify this by detecting bounding boxes and then segmenting each box independently. While effective, this sequential processing of proposals scales poorly with object density. \textbf{Bottom-Up} methods instead predict dense pixel-level embeddings that are aggregated post-inference. This family of approaches avoids the proposal bottleneck, offering a theoretically superior path toward parallelized, real-time decoding.}

\subsection{Background}
The instance segmentation approach discussed in this paper is introduced in \cite{prisadnikov2024simple}, where, in conjunction with a strong DINOv2 backbone \cite{oquab2023dinov2} and a carefully designed loss function~\cite{prisadnikov2024simple}, it achieves state-of-the-art results on the COCO dataset \cite{lin2014microsoft}. DINOv2 has already been successfully applied to segmentation~\cite{ren2024dino} and object detection~\cite{liu2024grounding,ren2024grounding}, making it a strong backbone model for a variety of tasks. This work extends the paradigm adopted in Painter \cite{wang2023images} and is conceptually related to DCME \cite{watanabe2018distance}. Painter encodes instances via dense RGB embeddings regressed per pixel, while DCME predicts per-pixel displacement vectors to instance centroids; sinusoidal positional encodings provide a more robust alternative to both approaches. In~\cite{scribano2025edgedeploy}, a basic CUDA-based decoder is presented to leverage GPU acceleration. However, this approach suffers from atomic contention and poor memory reuse, leading to sub-optimal performance.

\subsection{CUDA and GPU programming}

\red{Modern GPUs leverage the SIMD (Single Instruction Multiple Data) paradigm to provide massive parallel processing capabilities. Under the GPGPU framework, tasks are executed by transferring data to the device, launching a specialized kernel, and copying results back to the host.
Using the CUDA programming model, developers can explicitly manage this parallelism by configuring threads into a grid of three-dimensional blocks. This hierarchy allows for fine-grained control over resources: threads within the same block utilize high-speed, on-chip shared memory and barriers for synchronization. By leveraging shared memory as a low-latency cache for frequently accessed data, the overhead of global memory access is mitigated, significantly enhancing the overall performance of the implementation.}

\section{Methodology}

\blue{We adopt the encoding formulation proposed in~\cite{prisadnikov2024simple}.}
Let $M = \{M_k\}_{k=1}^{N}$ be the set of ground-truth instance masks. For each instance $k$, let $(x_c^k, y_c^k) \in \mathbb{R}^2$ denote the image-space coordinates of its centroid, \red{normalized in $[-1,1]$}, and let $\Omega_k \subset \{1,\dots,H_O\} \times \{1,\dots,W_O\}$ denote the set of \red{normalized} pixel locations belonging to mask $M_k$. We define the instance-segmentation target $T \in \mathbb{R}^{H_O \times W_O \times 4L}$ such that each spatial location encodes the sinusoidal positional encoding of the centroid of the instance it belongs to, or a void encoding for background pixels. Formally, for each location $(i,j)$:
\begin{equation}
\label{eq:is}
    T(i,j) =
    \begin{cases}
    \texttt{concat}(\gamma(x_c^k), \gamma(y_c^k)) & \text{if } (i,j) \in \Omega_k \\ %
    \mathbf{0} \in \mathbb{R}^{4L}
    & \text{otherwise.}
    \end{cases}
\end{equation}
where the positional encoding $\gamma(p) : \mathbb{R}\to\mathbb{R}^{2L}$, derived from \cite{vaswani2017attention}, is defined as:
\begin{equation}
    \gamma(p) = (sin(2^l\pi p),cos(2^l\pi p))_{l=0}^{L-1}
\end{equation}
$L$ defines the number of used harmonics; $L=4$ is used.

\subsection{Decoding Algorithm}
The algorithm to recover the predicted instance masks $\hat{M}$ from a noisy estimate $\hat{T}$ requires several steps. Let the network prediction at location $(i,j)$ be:
\begin{equation}
\hat T(i,j)=\texttt{concat}\left(\hat{x}(i,j),\hat{y}(i,j)\right)
\quad
\hat{x}(i,j),\hat{y}(i,j)\in\mathbb{R}^{2L}
\end{equation}

First, a coarse voting histogram $H\in\mathbb{R}^{(B_w +1)\times(B_h + 1)}$ is computed, with $\mathcal{B} = [1,...,B_w +1]\times[1,...,B_h +1]$ denoting the set of \textbf{bins} locations \red{(each representing a candidate instance centroid)}. Each output location $(i,j)$ casts soft votes for all histogram bins $(a,b)\in\mathcal{B}$ by comparing its predicted encodings to the positional encodings of the bin coordinates. The separable distance between output location $(i,j)$ and histogram bin $(a,b)$ in the encoded space is defined as $d_{ij}(a,b) = d_x(i,j,a)+d_y(i,j,b)$, with:
\begin{equation}
    \begin{gathered}
    d_x(i,j,a)= {||\hat x(i,j)-\gamma(a)||} \\
    d_y(i,j,b)= {||\hat y(i,j)-\gamma(b)||}
\end{gathered}
\end{equation}
The corresponding vote weight is given by the negative exponential of the distance:
\begin{equation}
    w_{ij}(a,b) = e^{-d_{ij}(a,b)}
    \label{eq:dist}
\end{equation}
The coarse voting histogram $H$ is then obtained by aggregating votes over the set of  participating  output locations $\mathcal{S}=[1,...,H_O]\times[1,...,W_O]$:
\begin{equation}
    H(a,b)=\sum_{(i,j)\in\mathcal{S}} w_{ij}(a,b)
    \label{eq:h}
\end{equation}
Local-maxima search is used to retrieve a set of $\hat K$ candidate mask centroids $\mathcal{C}=\{(a_k,b_k)\}_{k=1}^{\hat K}$, while simultaneously suppressing weak centroid candidates caused by prediction noise. Each element of $\mathcal{C}$ represents a hypothesized instance centroid in the discretized output coordinate space. Each output location $(i,j)$ contributes to the $m$-th mask, $M_m$, associated with the closest candidate centroid bin in the encoding space: %
\begin{equation}
\label{eq:argmin}
    m = \text{arg} \min_kd_{ij}(a_k, b_k)
\end{equation}
\begin{equation}
\label{eq:m}
    \hat M_m(i,j) = \begin{cases}
        1 \quad \text{if} \quad d_{ij}(a_m,b_m)<\tau\\
        0 \quad \text{otherwise}
    \end{cases}
\end{equation}

\subsection{Decoding Implementation}
In this section, we describe our implementation of the decoding phase. We exploit the CUDA ecosystem to offload the computation on the GPU. 
First, we precompute the \textit{palettes} for the encoded bin coordinates in $x$ and $y$ directions, denoted as $P_w$ and $P_h$ respectively. Where:
\begin{equation}
    P_n = \left \{ \gamma \left(-1 + \frac{2i}{n+1}\right) \Bigg |  i=1,2,\dots,n\right\}
\end{equation}
denotes the vector of precomputed $\gamma(\cdot)$  evaluated at evenly spaced candidate centroid locations corresponding to histogram bins. Unlike the original implementation, we do not use an explicit void encoding for background pixels; pixels not associated with any valid instance are implicitly treated as background, and the centroid voting histogram therefore consists of $B_w\times B_h$ bins. \red{$P_h$ and $P_w$ are stored as model parameters preloaded in memory and can be accessed without recomputing or memory copying.}\\

\paragraph{Histogram accumulation.}

First, we compute $H$ from Equation~\ref{eq:h}. The naive implementation~\cite{scribano2025edgedeploy} assigns one $((i,j),(a,b))$ tuple to each thread. The thread explicitly computes $w_{ij}(a,b)$ for the assigned pixel-bin pair and adds the result to the global histogram bin $H(a,b)$. Since many threads vote for the same histogram bin concurrently, this accumulation requires \texttt{atomicAdd} operations on global memory. As a result, updates to popular bins become serialized, creating severe contention, limiting memory throughput, and reducing parallel efficiency. To overcome this limitation, we approach the problem by relying on a decomposition of $H$ and on the use of the optimized \textit{cuBLAS}~\footnote{https://developer.nvidia.com/cublas} library. The sum over $\mathcal{S}$ in \Cref{eq:h} admits the following decomposition by factorizing the exponential of the sum in $w_{i,j}(a,b)$ as product of exponentials $w_{i,j}(a,b)=E_x(i,j,a)E_y(i,j,b)$ where:
\begin{equation}
    \begin{gathered}
    E_x(i,j,a)= e^{-||\hat x(i,j)-P_w(a)||} \\
    E_y(i,j,b)= e^{-||\hat y(i,j)-P_h(b)||}
\end{gathered}
\end{equation}
Therefore, \Cref{eq:h} decomposes as:
\begin{equation}
\label{eq:new_h}
    H(a,b)= E_x^\top E_y
\end{equation}
Delegating the computation of $E_x$ and $E_y$ to specialized kernels, this formulation eliminates the need for an accumulator and atomic operations. The two kernels are launched with $H_O*W_O*a$ threads ($H_O*W_O*b$ for $E_y$ matrix), so each thread computes a single element of the corresponding matrix. Once $E_x$ and $E_y$ are computed, $H$ is obtained using the optimized \textit{cuCBLAS} function \texttt{cublasSgemmStrided} to implementing Equation~\ref{eq:new_h}.\\

\paragraph{Local Maxima Search.}
The computed $H$ is a noisy estimate of candidate mask centroids (\Cref{fig:decode_b}). The Local Maxima Search stage produces a sparse binary indicator map $\overline{H}$ marking which \red{bins} are selected as candidate centroids (\Cref{fig:decode_c}). Conceptually, the kernel assigns one CUDA thread to each histogram cell $(x,y)$. Each thread loads the center value $H(x,y)$ and tests whether it is a strict local maximum within its 8-connected neighborhood. If this condition is satisfied and the value exceeds a predefined threshold, a binary peak indicator is written to $\overline{H}(x,y)$.\\

\paragraph{Mask Aggregation.}

The final stage assigns each output pixel to an instance mask according to Equation~\ref{eq:m}. The kernel is launched with one thread per output pixel $(i,j)$. Given the binary centroid map $\overline{H}$ and the affinity matrices $E_x$ and $E_y$, each thread scans all active centroid bins $(a,b)$ such that $\overline{H}(a,b)=1$. For each candidate, it computes the combined affinity $E_x(i,j,a) * E_y(i,j,b)$ and selects the centroid with the highest value among the candidates that satisfy the affinity threshold $e^{-\tau}$. Before scanning the centroids, each thread loads the affinity values associated with its pixel into shared memory. This avoids repeatedly reading the same $E_x$ and $E_y$ entries from global memory while evaluating multiple candidate centroids. The output memory is preallocated for all $B_w \times B_h$ possible masks, one per histogram bin. Once the best centroid $(a_m,b_m)$ is selected, the thread sets $\hat{M}_m(i,j)=1$ in the corresponding output mask. The kernel also maintains a counter vector $C$, where $C_m$ stores the number of pixels assigned to mask $m$. Since multiple threads may assign pixels to the same mask, updating this counter requires an \texttt{atomicAdd}. In practice, this atomic operation is limited to one update per assigned pixel, while the main centroid search is performed locally by each thread.

\noindent Compared with a naive implementation \cite{scribano2025edgedeploy} that launches one thread per tuple $(i,j,m)$, this design reduces memory traffic by evaluating all candidate masks for a pixel within a single thread and reusing the pixel-specific affinity values from shared memory. This avoids repeatedly loading the same affinity vectors for each candidate mask and improves the efficiency of the aggregation stage.

\section{Results and Discussion}

In this experimental section, we discuss the performance of our CUDA-accelerated implementation of the decoding algorithm.
\blue{We are interested in both system and task metrics. In particular, we measure the inference time of the decoder (Inference Time), the End-to-End inference time of the model (End-to-End Inference Time), the memory footprint, and the Panoptic Quality (PQ)~\cite{kirillov2019panoptic}.}
First, we identify a latency/performance tradeoff varying the number of histogram bins, while also validating the task performances against the PyTorch reference implementation\cite{prisadnikov2024simple}. Then we assess the latency and memory footprint at different input resolutions. Finally, we break down the End-to-End inference performance, including model inference time, to showcase the significant impact of the decoding stage.

\subsection{Experimental setup}
\begin{figure}
    \centering
    \includegraphics[width=0.7\linewidth]{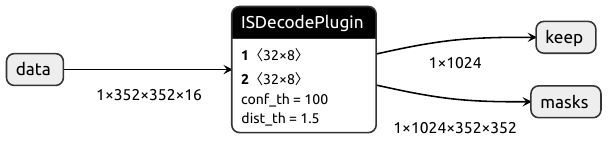}
    \caption{Dummy \texttt{ONNX} model used to profile TensorRT execution.}
    \label{fig:onnx}
\end{figure}

\red{Latency and memory consumption are profiled on an NVIDIA Jetson Orin Nano, using TensorRT for inference acceleration. The decoder is implemented as a TensorRT plugin via the \texttt{IPluginV3} interface. To evaluate the decoder in isolation from the rest of the model, we designed a minimal ONNX graph (\Cref{fig:onnx}) which is compiled into a TensorRT engine. We leverage the \texttt{trtexec} utility to measure execution time, omitting data transfer latencies\footnote{trtexec --loadEngine=instseg.trt --noDataTransfers --useSpinWait
--iterations=100 --avgRuns=100 --warmUp=10 --staticPlugins=ISDecodeIPluginV3.so}.}

\red{The memory footprint is assessed by implementing a custom CUDA allocator within the TensorRT environment to track peak memory usage during inference. Task performance (\Cref{tab:bins}) is quantified using the Panoptic Quality (PQ) metric, following the protocol established in \cite{prisadnikov2024simple}. To isolate the impact of our optimizations, we evaluate PQ across different decoder implementations while keeping the rest of the model architecture constant.}

\subsection{Impact of the number of bins}

\begin{table}
\centering
\caption{Comparison of execution time and memory footprint of different decoder implementations.}
\begin{tabularx}{\columnwidth}{lYYYYYYYY}%
\toprule
\multirow{2}{*}{\textbf{Bins}} & \multicolumn{2}{c}{\textbf{PQ}} & \multicolumn{2}{c}{\textbf{Infer time (ms)}} & \multicolumn{2}{c}{\textbf{Infer time st.dev}} & \multicolumn{2}{c}{\textbf{Memory (MB})} \\ \cmidrule{2-9} 
    & \textbf{Torch}      & \textbf{TRT}       & \textbf{Torch}          & \textbf{TRT}           & \textbf{Torch}             & \textbf{TRT}               & \textbf{Torch}      & \textbf{TRT}        \\ \cmidrule(r){1-1} \cmidrule(lr){2-3} \cmidrule(lr){4-5} \cmidrule(lr){6-7} \cmidrule(lr){8-9}%
$(8\times8)$                    & 26.68      & 22.57     & 56.44          & 2.13          & 0.16              & 0.001             & 64.55      & 47.58      \\ 
$(16\times 16)$                    & 43.72      & 34.91     & 74.91          & 6.13          & 0.20              & 0.003             & 77.54      & 150.67     \\ 
$\mathbf{(32\times 32)}$                    & 51.32      & 50.78     & 196.18         & 30.66         & 0.54              & 0.01              & 121.92     & 547.18     \\
$(64\times 64)$                    & 52.26      & 52.08     & 493.76         & 101.37        & 27.01             & 0.017             & 218.25     & 2101.47    \\
$(80\times 80)$                    & 52.34      & 52.14     & 687.99         & 144.00        & 42.18             & 0.026             & 272.25     & 3259.25    \\ \bottomrule
\end{tabularx}
\label{tab:bins}
\end{table}

\red{Decoder efficiency and task accuracy are primarily influenced by the histogram bin resolution. We assess this trade-off by comparing our CUDA implementation against the original PyTorch reference~\cite{prisadnikov2024simple}. Results in Table~\ref{tab:bins} demonstrate that our decoder is consistently faster and exhibits superior timing stability, which is essential for predictable real-time performance in safety-critical applications. This stability stems from our use of static memory pre-allocation during the initialization phase. Our analysis reveals that PQ gains follow a trend of diminishing returns, plateauing at a resolution of $32 \times 32$ bins. Given that higher resolutions increase latency without significant quality improvements, we adopt the $32 \times 32$ configuration for all subsequent evaluations.}

\subsection{Impact of the feature resolution}

\begin{table}
\caption{Inference time and PQ performance as a function of bin resolution for baseline and optimized decoders.}
\centering
\begin{tabularx}{\columnwidth}{lYYYY}
\midrule
\textbf{Input Res.}      & \textbf{Precision} & \textbf{End-to-end Infer - Torch (ms)} & \textbf{End-to-end Infer - Our (ms)} & \textbf{speedup \%} \\ \midrule
\multirow{3}{*}{$280\times280$} & \textbf{FP32}      & 225.61                                 & 154.56                      & 31.49               \\ \cmidrule{2-5} 
                         & \textbf{FP16}      & 138.40                                 & 67.35                       & 51.34               \\ \cmidrule{2-5} 
                         & \textbf{INT8}      & \textbf{129.12}                        & \textbf{58.07}              & \textbf{55.03}      \\ \midrule
\multirow{3}{*}{$336\times336$} & \textbf{FP32}      & 319.17                                 & 227.09                      & 28.85               \\ \cmidrule{2-5} 
                         & \textbf{FP16}      & 183.27                                 & 91.19                       & 50.24               \\ \cmidrule{2-5} 
                         & \textbf{INT8}      & 171.15                                 & 79.07                       & 53.80               \\ \midrule
\multirow{3}{*}{$448\times448$} & \textbf{FP32}      & 595.24                                 & 482.33                      & 18.97               \\ \cmidrule{2-5} 
                         & \textbf{FP16}      & 284.85                                 & 171.94                      & 39.64               \\ \cmidrule{2-5} 
                         & \textbf{INT8}      & 260.71                                 & 147.80                      & 43.31               \\ \midrule
\multirow{3}{*}{$616\times616$} & \textbf{FP32}      & \textit{1332.53}                       & \textit{1167.01}            & \textit{12.42}      \\ \cmidrule{2-5} 
                         & \textbf{FP16}      & 507.73                                 & 342.21                      & 32.60               \\ \cmidrule{2-5} 
                         & \textbf{INT8}      & 465.52                                 & 300.00                      & 35.56               \\ \midrule
\end{tabularx}
\label{tab:e2e_times}
\end{table}

We further evaluate the scalability of our decoder with respect to the encoding space resolution $\hat T \in \mathbb{R}^{H_O\times W_O\times 4L}$. 
Since the spatial dimensions $H_O, W_O$ scale with the model input resolution, we measure inference time for different feature-map sizes using a fixed $32\times32$ bin configuration (Figure~\ref{fig:res}). The proposed implementation consistently outperforms the PyTorch reference and exhibits better scaling behavior as the feature-map resolution increases. These results indicate that the optimized kernels effectively limit the additional computational cost associated with higher-resolution inputs.

\subsection{End-to-End Latency Breakdown}

To evaluate system-level performance, we integrate the proposed decoder into the model from \cite{prisadnikov2024simple}. The network uses a DINOv2 \cite{oquab2023dinov2} (ViT-L \cite{dosovitskiy2020image}) backbone and four transposed convolutions for feature upsampling. After export to ONNX, the model is optimized with TensorRT and benchmarked at different precision levels. This configuration enables a realistic assessment of the decoder's contribution to end-to-end inference performance. We report the End-to-End latency at different combinations of input resolution ($280\times280$, $336\times336$, $448\times448$ and $616\times616$) and numerical precision (FP32, FP16, and INT8). %

\begin{figure}[t]
    \centering
    
    \begin{subfigure}[t]{0.48\linewidth}
        \centering
        \includegraphics[width=\linewidth]{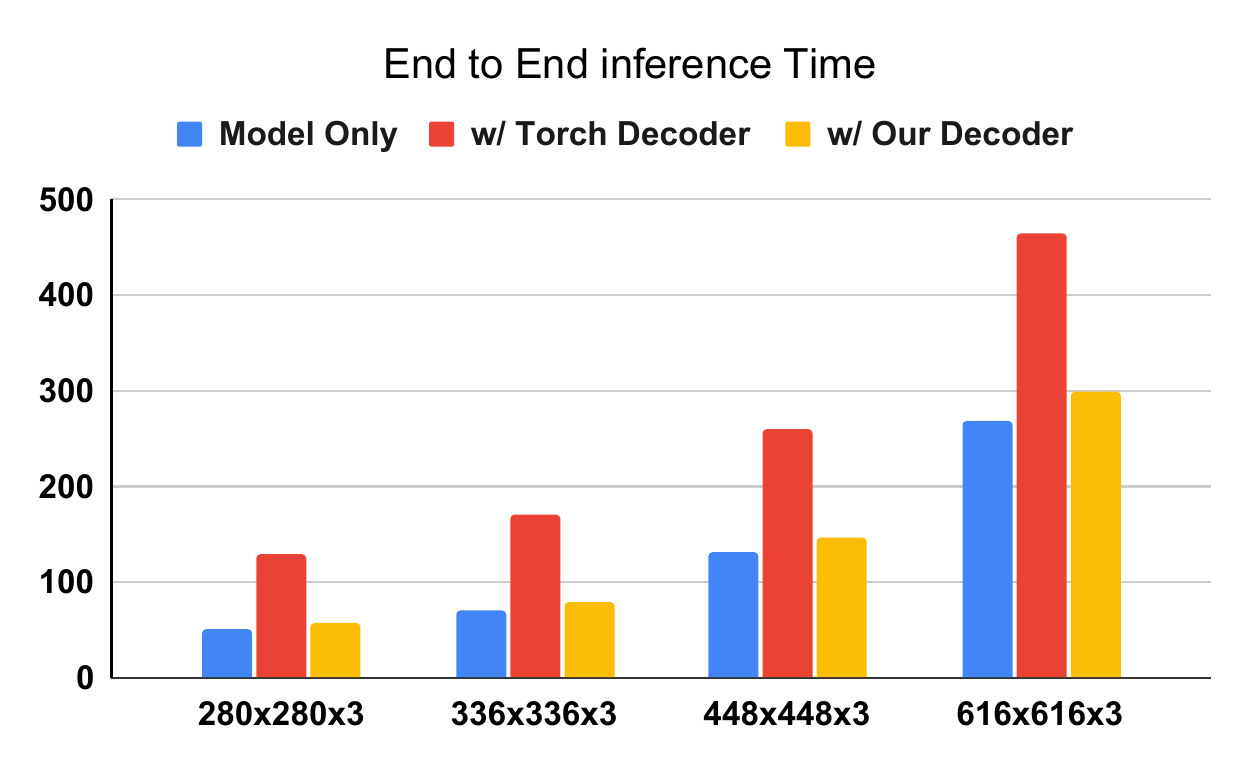}
        \caption{End-to-End inference latency comparison across varying resolutions at INT8 precision.}
        \label{fig:e2e}
    \end{subfigure}
    \hfill
    \begin{subfigure}[t]{0.48\linewidth}
        \centering
        \includegraphics[trim={25 25 0 0},clip, width=\linewidth]{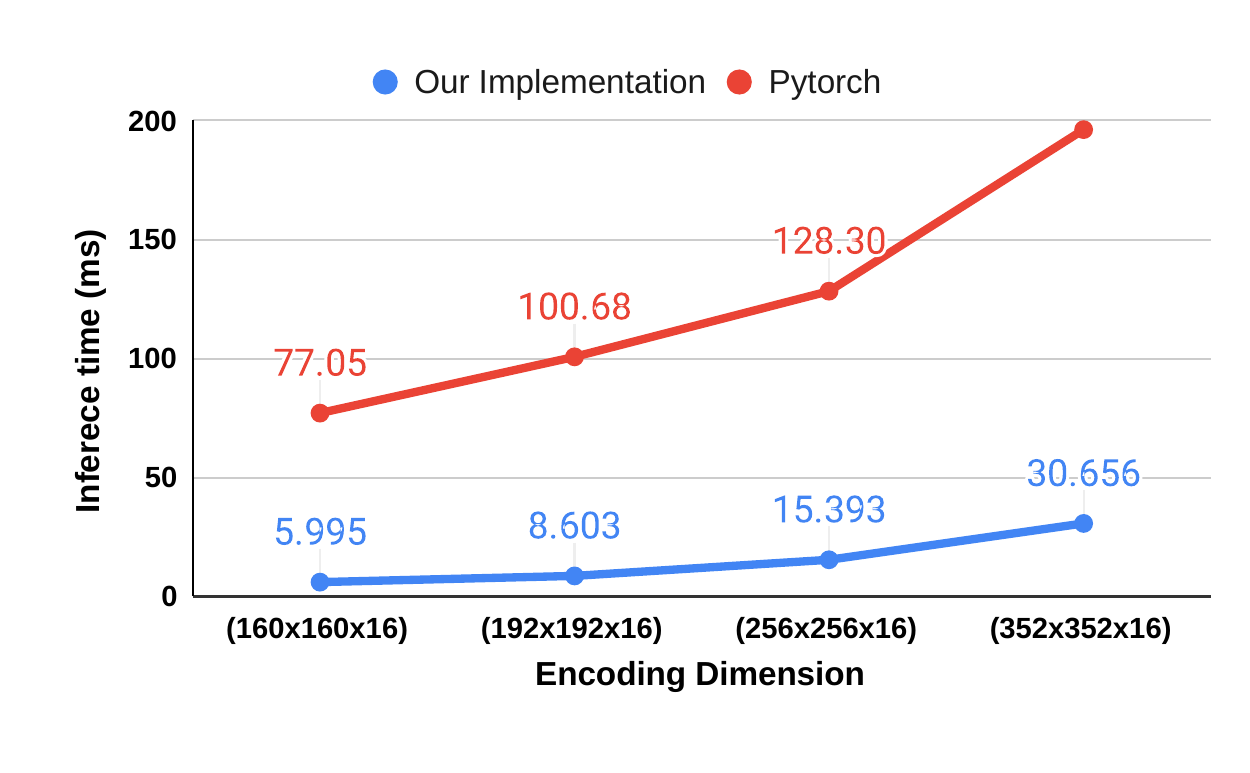}
        \caption{Impact of encoding resolution on latency: PyTorch baseline vs. optimized CUDA implementation (fixed $32\times 32$ bins, $L=4$).}
        \label{fig:res}
    \end{subfigure}
    
    \caption{Latency analysis across implementations and resolutions.}
    \label{fig:combined}
\end{figure}
\red{The end-to-end performance comparison is reported in Table~\ref{tab:e2e_times}. To isolate the impact of the decoder, both pipelines use the same TensorRT-optimized backbone and intermediate model components; the only difference is the decoding stage, which is implemented either using the original PyTorch decoder or the proposed CUDA implementation. Under these conditions, the proposed decoder consistently reduces end-to-end latency, achieving speedups of up to 55\% in INT8 mode at a resolution of $280\times280$. The comparison in Figure~\ref{fig:e2e} highlights that decoding is a non-trivial bottleneck and can dominate the End-to-End execution time when relying on a suboptimal implementation. }%

\subsection{Kernel Execution Profiling}

\begin{table}
\caption{Execution time of each kernel}
\centering
\begin{tabular}{@{}llcccc@{}}
\toprule
                                    & \multicolumn{1}{c}{\textbf{Operation}} & \textbf{Kernel}         & \textbf{Exec (ms)} & \textbf{Exec std (ms)} & \textbf{time \%} \\ \midrule
\multirow{6}{*}{\textbf{Optim.}}     & \multirow{4}{*}{HistAcc}        & \texttt{compute ex kernel}       & \textit{6.47}    & 0.001  & 21.09             \\
                                    &                                        & \texttt{compute ey kernel}       & 6.47        & 0.001       & 21.09             \\
                                    &                                        & \texttt{cublasSgemmStrided}      & 0.75          & 0.002     & 2.44              \\
                                    &                                        & { \textit{total}}                & 13.68         &           & 44.62             \\ \cmidrule(l){2-6} 
                                    & LMS                    & \texttt{local\_maxima}           & 0.010       & 0.000       & 0.03            \\ \cmidrule(l){2-6} 
                                    & MaskAgg                       & \texttt{mask\_aggregation}      & 6.88        & 0.007       & 22.44             \\ \toprule
\multirow{3}{*}{\textbf{Naive}} & HistAcc                        & \texttt{fused\_vote\_kernel}     & 77.99       & 0.003       & 73.41             \\ \cmidrule(l){2-6} 
                                    & LMS                   & \texttt{local\_maxima\_nai}      & 0.008      & 0.000        & 0.01            \\ \cmidrule(l){2-6} 
                                    & MaskAgg                       & \texttt{mask\_aggregation\_nai} & 21.26       & 0.003       & 20.02             \\ \bottomrule
\end{tabular}
\label{tab:perc}
\end{table}

\red{We further break down the performance profile of our implementation by analyzing individual kernel latencies (Table~\ref{tab:perc}). The \textit{Histogram accumulation} (HistAcc) kernel is the most significant contributor to total latency, followed by \textit{Mask Aggregation}  (Mask Agg), while \texttt{cublasSgemmStrided} and \textit{Local Maxima Search} (LMS) have a marginal impact. The implementation demonstrates high timing stability, which is vital for safety-critical applications. The minor execution variance in the \textit{Mask Aggregation} and \texttt{cublasSgemmStrided} kernels stems from atomic-induced memory contention. Our design choices in the Optimized implementation have significantly curtailed raw GPU compute time with respect to the Naive implementation, shifting the performance bottleneck toward memory operations, which now account for 33\% of the total decoder inference time.}

A comparison with the naive implementation~\cite{scribano2025edgedeploy} highlights that the decomposition of $H$ and the use of \textit{cuBLAS} are the primary contributors to the improvement in decoder execution time. Additionally, the optimization of the \textit{Mask Aggregation} kernel provides a substantial benefit, achieving an approximately 6-fold speedup.
\blue{The naive and optimized kernels are mutually incompatible due to differing data encoding schemes in their respective preceding pipeline stages.} In summary, our optimizations drastically reduce the decoder execution time, which in turn lowers the End-to-End latency of the entire model, as the decoder represents a significant portion of the overall pipeline.

\section{Conclusion}

\red{In this work, we introduced a high-performance GPU-accelerated decoder for instance segmentation that achieves substantial speedups over existing implementations. Our results demonstrate that optimizing traditionally CPU-bound auxiliary stages is as critical to reducing End-to-End latency as refining the core model architecture. By outperforming baseline GPU implementations, our design validates the necessity of hardware-aware decoding. Future research will focus on optimizing memory management—which currently accounts for 33\% of the remaining latency—by transitioning from framework-level handling to custom, low-level memory orchestration.}

\begin{credits}
\subsubsection{\ackname} This research was partially funded by the dAIedge project (HORIZON-CL4-2022-HUMAN-02-02, Grant Agreement Number: 101120726) and the Ministry of Education and Science of Bulgaria (support for INSAIT, part of the Bulgarian National Roadmap for Research Infrastructure).

\subsubsection{\discintname}
The authors have no competing interests to declare that are
relevant to the content of this article.
\end{credits}
\bibliographystyle{splncs04}
\bibliography{main}
\end{document}